\documentclass[letterpaper, 10 pt, conference]{ieeeconf}  

\usepackage{amsmath}
\usepackage{amssymb}
\usepackage{xparse}
\usepackage{xspace}
\usepackage{url}
\usepackage{graphicx}
\usepackage{subfig}
\usepackage{mathtools}
\usepackage{booktabs}
\usepackage{xcolor}
\usepackage{mathtools}
\usepackage{float}
\usepackage{mciteplus}

\def\andothers{et al.\,}
\def\figname{Fig.\,}

\IEEEoverridecommandlockouts
\graphicspath{./images}

\title{Markerless Visual Robot Programming by Demonstration}

\graphicspath{{images/}}

\author{Raphael Memmesheimer$^{1}$, Ivanna Mykhalchyshyna$^{1}$, Viktor Seib$^{1}$, Nick Theisen$^{1}$ and Dietrich Paulus$^{1}$%
\thanks{$^{1}$Active Vision Group, Institute for Computational Visualistics,
        University of Koblenz-Landau, 56070 Koblenz, Germany
        {\tt\small \{raphael, ivannamyckhal, vseib, nicktheisen, paulus\}@uni-koblenz.de}}%
}

\begin{document}

\maketitle
\thispagestyle{empty}
\pagestyle{empty}

\begin{abstract}
    In this paper we present an approach for learning to imitate human behavior
    on a semantic level by markerless visual observation. We analyze a set of spatial 
    constraints on human pose data extracted using convolutional pose machines and 
    object informations extracted from 2D image sequences.
    A scene analysis, based on an ontology of objects and affordances, is combined with continuous human pose estimation and spatial object relations.
    Using a set of constraints we associate the observed human actions with a set of executable robot commands.
    We demonstrate our approach in a kitchen task, where the robot learns to prepare a meal.
\end{abstract}

\section{Introduction}
\label{sec:introduction}
Programming by Demonstration (PbD) \cite{billard2008robot}
is an alternative method for teaching robot tasks where no expert programmer is needed.
Instead, an expert demonstrator shows
a robot a task to imitate. Robots could be programmed textually \cite{gruver1984industrial}, 
by graphical user interfaces \cite{thomas2013new, mackenzie1998evaluating}, touch 
guidance \cite{schneider2010robot}, teleoperation or actual demonstration of tasks \cite{7029975, pastor2009learning}.
We believe that common programming methods for robots will be to complex to handle
specialized tasks. 
In the future it can be expected that robots are employed in almost all fields in human life.
This raises the need for alternative programming methods for complex tasks.

PbD is already used by many industrial service robots \cite{friedrich1996robot, aleotti2003toward}.
Other approaches are mostly focusing on adding markers \cite{ALEOTTI2006409, 7029975, pastor2009learning}
or external sensors to the demonstrator \cite{calinon2007active}, usually interfering with the demonstrator during
the task.
We propose an approach that does not rely on changes of the environment 
and does not rely on additional sensors except a common RGB camera.

Current robotic systems
that lack a certain desired behavior, commonly need an expert programmer to add the missing functionality.
Contrary, we introduce an approach related to programming robots by visual demonstration \cite{azad2009visual}
that can be applied by common users.
Provided a basic scene understanding, the robot observes a person demonstrating a task and is then able
to reproduce the observed action sequence using its semantic knowledge base.

\begin{figure}[!t] 
  \includegraphics[width=\linewidth]{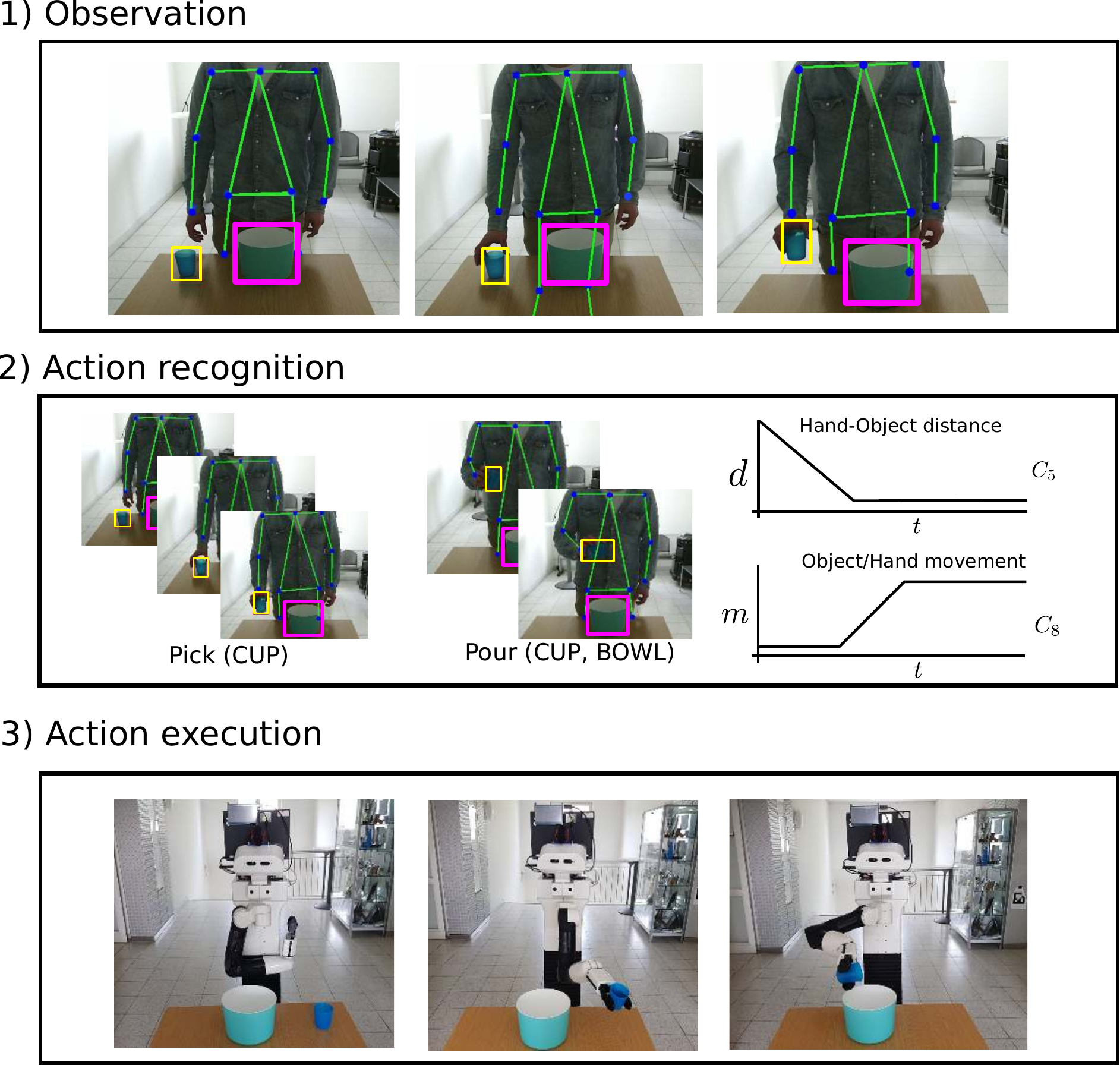}
  \caption{Approach overview for extracting action informations from 2D image sequences in order to
  execute them on a mobile robot. 
  Exemplary object detections (yellow, pink) and human pose estimates (green) are observed. 
  Actions are recognized using a set of constraints. 
  For replicating the observed actions we used two mobile robots equipped with an arm.
  } 
  \label{fig:approach}
\end{figure}

We image different use cases for our approach.
The observed action
can be associated with a command given by natural language, where parameters
are exchangeable due to the semantic representation.
Further, robots observing (and interpreting) tasks can create a task representation.
This representation can be transfered to other robots \cite{figueroa2016learning}
in order to execute the task without ever seeing it.
Trajectory level learning combined with a task level representation learning could
deprecate currently common programming methods for robot and enable non expert 
programmer to teach new tasks from ground up \cite{caccavale2018kinesthetic}.

The core contribution is an approach for markerless action recognition based on
Convolutional Pose Machines (CPM) \cite{DBLP:journals/corr/WeiRKS16},
 object observations \cite{DBLP:journals/corr/RedmonDGF15} and continuous spatial relations. 
We show that the actions are executable on a robot that is able to execute a 
set of common actions. The initial scene analysis allows semantic reasoning in case
the required object is not present.
Further, this allows executing the same action sequence with different objects 
which is a major benefit over action sequencing approaches that rely on 
positional data only. 
Even so we are demonstrating our approach on 2D observations, the formulations
are also adaptable in 3D.
\figname \ref{fig:approach} gives an overview of our approach.

This paper is structured as follows. In Section \ref{sec:related_work} we discuss
similar approaches. Section \ref{sec:approach} describes our approach in
depth. Experimental results are shown in Section \ref{sec:experiments}.
Section \ref{sec:conclusion} concludes the paper with an outlook.


\section{Related Work}
\label{sec:related_work}
Many approaches for programming robot systems have been proposed in the past. Most
common are approaches for guidance by force sensors \cite{seidel2014model,figueroa2016learning}, teleoperation or sensory 
based approaches \cite{vakanski2017robot, 7029975, ALEOTTI2006409}. 
Sensor based approaches, like motion capturing camera systems are relying on 
active changes of the environment. Most commonly the changes are done by attaching either reflective
or non-reflective markers to the demonstrator and/or the interacting objects \cite{7029975, ALEOTTI2006409}. 
Most common are teaching by guidance approaches \cite{seidel2014model} 
that allow the guidance of a robot arm through force sensors.
These approaches are subsumed under the term learning by demonstration.
It usually describes 
approaches relying on a demonstrator moving a robot arm to perform a task.
Using force torque sensors the arm is able to recognize the forces applied by a demonstrator.
Throughout the movements, the joint angles are recorded and used for a later
replay. 

W\"achter \andothers \cite{7029975} have shown an approach for action sequencing
that analyses the demonstrated task by motion capture measurements and have executed 
the observed behavior by a humanoid robot.  
More advanced approaches take the pose of the robot arm and combine object estimates
to act not just on a trajectory level, but associate the interacting objects. 
Koskinopoulo \andothers \cite{7451734} formulate a latent representation of task observations
demonstrated by a human and employ them for human robot collaborative tasks.
Magnanimo \andothers \cite{6926339} proposed an approach to recognize tasks executed by a human
and predict next actions and objects to manipulate using a Dynamic Bayesian Network.
Schneider \andothers \cite{schneider2010robot} presented integrated object estimates
and adapt touch guided trajectories to new object positions with Gaussian processes.

As we have pointed out, most of the approaches have in common to actively 
attach sensors or markers on the demonstrator. Only a small amount of approaches
\cite{azad2009visual} 
deal with robotic systems that only employ the onboard sensor setup as an observer.
Further, approaches like \cite{alexandrova2015roboflow} \cite{alexandrova2014robot} 
support the users by providing graphical user interfaces for programming and 
giving visual feedback during the programming procedure. Other approaches
allow to program a robot by natural language \cite{forbes2015robot} with a low-level set of commands guiding 
the arm movements.




\section{Approach}
\label{sec:approach}



\begin{figure*}[t!]
    \centering
    \begin{minipage}{.6\linewidth}
        \centering
        \includegraphics[width=\linewidth]{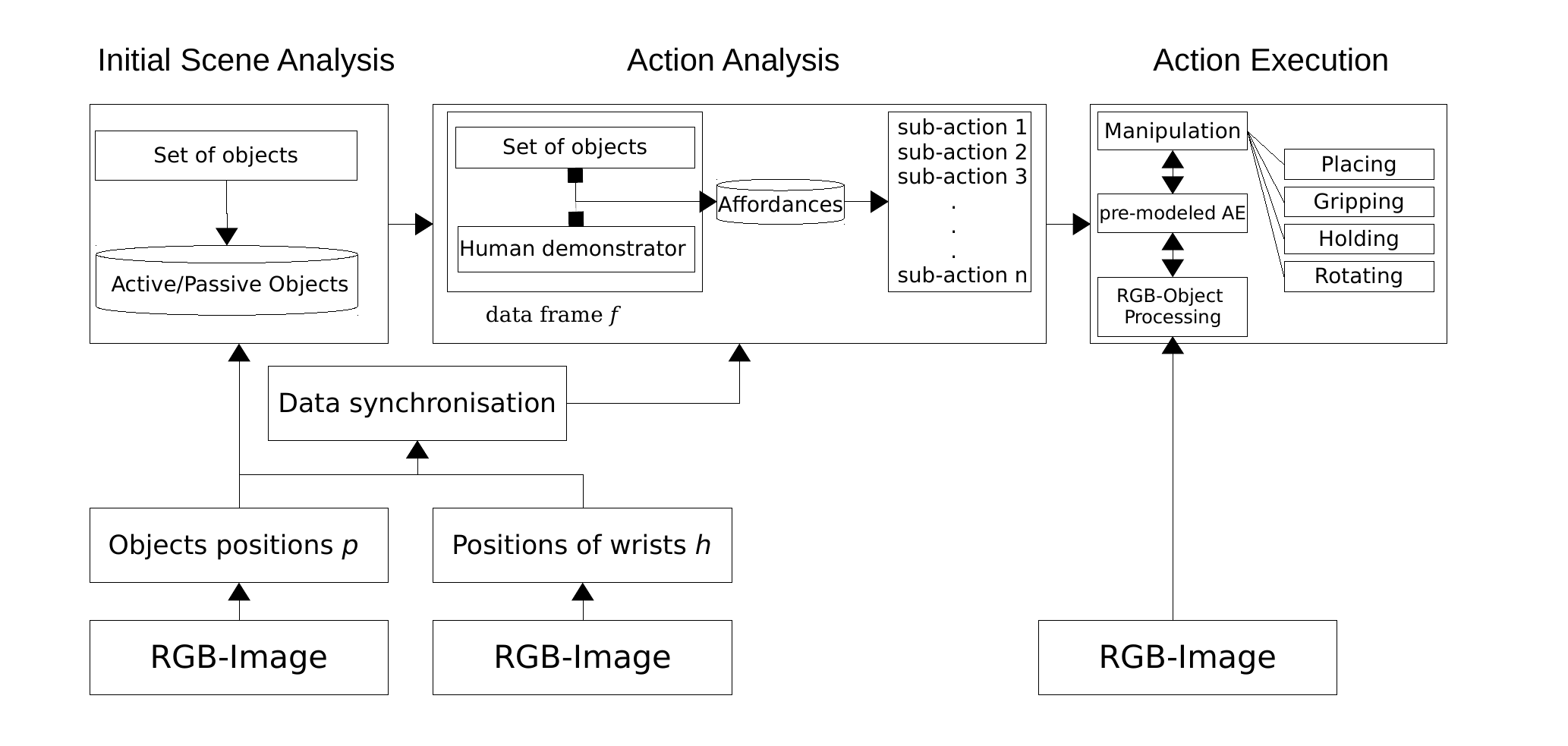}
        \caption{Overview of our approach in action sequencing}
        \label{fig:overview}
    \end{minipage}%
    \begin{minipage}{.4\linewidth}
        \centering

        \includegraphics[width=\linewidth]{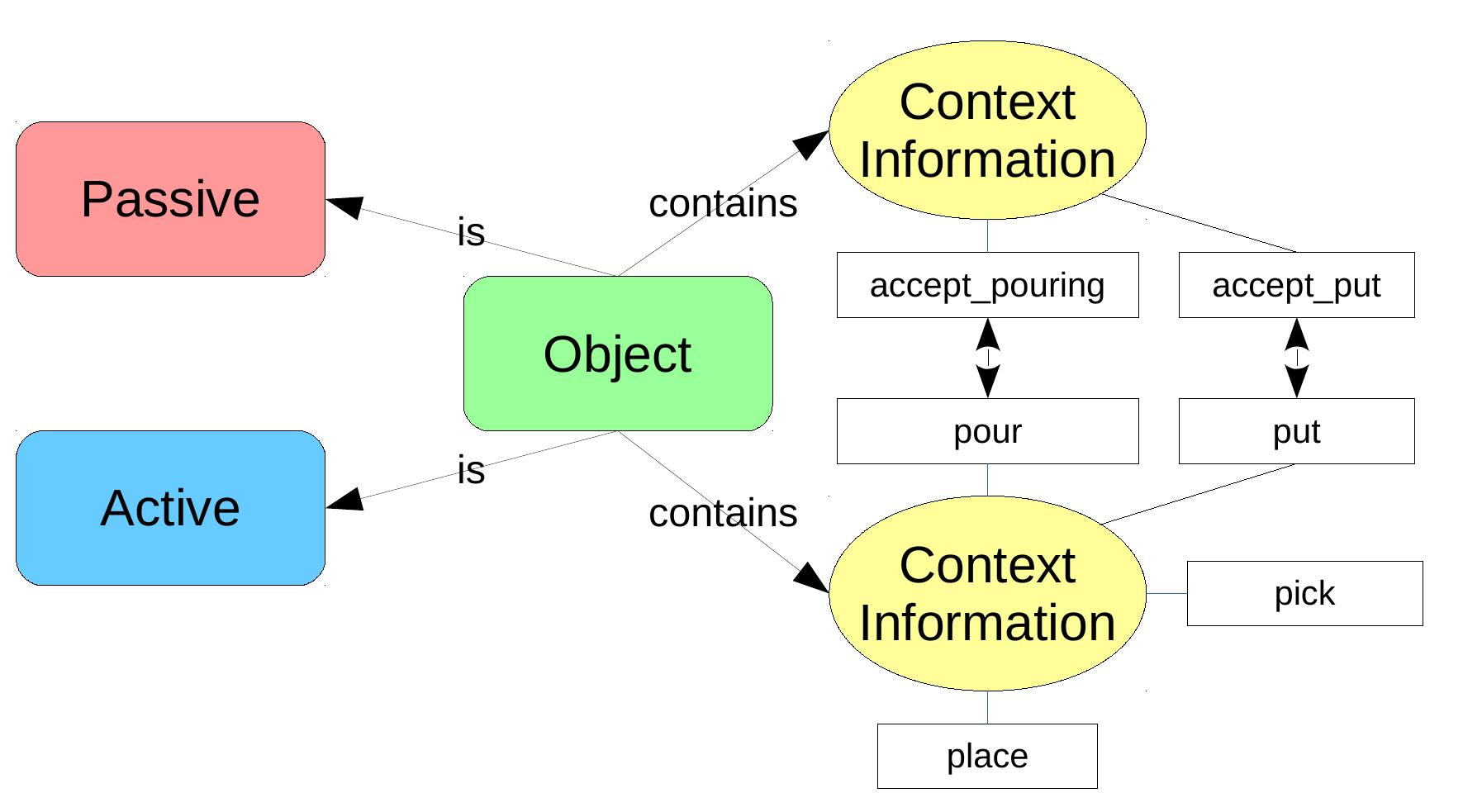}
        \caption{Ontology for Object Affordances in manipulation task}
        \label{fig:ontology}
    \end{minipage}
\end{figure*}

In this section, we describe our proposed approach in detail.
\figname \ref{fig:overview} gives an overview over our approach.
First, the robot analyzes the scene from its point of view using only on-board sensors.
The goal is to detect objects and a demonstrator.
Then, the robot starts observing the joints with special focus on the hands and object's positions in the scene.
Note that our approach requires prior knowledge about the objects used.
For this purpose, the scene overview is forwarded to a neural network (YOLO \cite{DBLP:journals/corr/RedmonDGF15}), which assigns a class label to every detected object.
Based on RGB images only, spatial relations between hands and objects, including semantic relations between objects and actions, are be segmented.
As we work only with 2D data we need an additional mechanism which allows representing complex contextual knowledge for objects and possible actions.
For this purpose we use a semantic knowledge base, where object affordances are modeled with an ontology.
The ontology represents semantic relations between objects and possible actions as depicted in \figname \ref{fig:ontology}. 
Finally, the robot matches the sequence of observed sub-actions with a set of predefined reusable actions. \\

\subsection{Initial Scene Analysis}
\label{sub:initial_analysis}

We assume that the robot is facing a table where a person is about to start 
a demonstration. First, the robot gets an initial view of the scene by checking
if a human demonstrator is visible. Then the objects on the table are analyzed. 
During the initializing step we first assign every object $o$ from the set of trained objects $o \in O$, where $\left\vert{O}\right\vert = N$ is the number of trained objects, 
with two possible classes - \textit{active} and \textit{passive} for manipulable (cup, watering can) and not-manipulable (bowl, plant) objects.
Note, that only objects that can be safely manipulated with one hand are manipulable in this context, since the experiments are performed on a robot with only one arm.
We store the initial positions  of \textit{passive} objects and both, position and approximated local changes of bounding boxes of \textit{active} objects (see \figname \ref{fig:approach}). We use the local changes in order to determine complex actions such as pouring.  \\
The object ontology is used in the current and next step of the proposed approach to build constraints for observation-action mapping.

We denote the observed object $o_k, k \in \{0,1,...,N\}$ and it's position as the estimated centroid point $p_{c} \in \mathbb{R}^2$. In every RGB-frame we estimate hand positions and express them as $h$, with $h \in \mathbb{R}^2$.
Data extracted from RGB-images is synchronized over time and stored in a common data frame $f$.
Moreover, we track local changes of the object in order to detect a pouring action.
For this purpose, in every frame $f_i$ we calculate a vector $\vec{v}$ using object bounding box coordinates
\begin{align}\label{eq:local_changes}
\vec{v} = 
\begin{bmatrix}
x \\ y
\end{bmatrix} = 
\begin{bmatrix}
x_b \\ y_b
\end{bmatrix}
- 
\begin{bmatrix}
x_t \\ y_t
\end{bmatrix},
\end{align}
where  $x_t, y_t$ are the pixel coordinates of the top left corner and $x_b, y_b$ are the pixel coordinates of the bottom right corner of the object's bounding box in 2D image space. 
From the initial scene analysis we store the initial centroid of every detected object as $p_{c_kinit}$.
Further, for look up over the object ontology and assigning object $o$ to \textit{active\_object} or \textit{passive\_object} class we define a function $g_m(o)$ as
\[ 
	g_m(o) = \left\{
			\begin{array}{@{}l@{\thinspace}l}
				1, &\text{if $o$ is manipulable} \\
				0, &\text{otherwise.} \\
			\end{array}
			\right.
\]
Furthermore, we present a set with affordances $A$ and a function $g_l(o)$ for retrieving all possible affordances for a given object $o$ from the knowledge base:
\begin{equation} \label{eq:ga}
 		g_l(o) = A_o \subseteq A.
\end{equation} 
A third function $g_a(o,a)$ indicates, whether a specific affordance $a$ is a valid affordance of object $o$: 
 \[ 
 g_a(o,a) = \left\{
 \begin{array}{@{}l@{\thinspace}l}
 1, &\text{if $a \in A_o$ } \\
 0, &\text{otherwise.} \\
 \end{array}
 \right.
 \]
 
 We additionally store the table bounding box $T$ using the extracted top left $t_{tl}$ and bottom right $t_{br}$ corners of the table plane in 2D pixel space during initial scene analysis:
 \begin{align}\label{eq:table}
 T = \{ 
	t_{tl},
	t_{br}
 \}.
 \end{align}

\subsection{Action Analysis}
\label{sub:action_analysis}

After the scene initialization the human demonstrator starts the manipulation task.
During the demonstration we estimate the human's joint positions by CPM \cite{DBLP:journals/corr/WeiRKS16}, which provides human skeleton data containing 18 keypoints as it is shown in \figname \ref{fig:approach}.
The wrist positions are retrieved in image coordinates using the RGB camera. 
Moreover, we extract a list with detected objects from RGB-data and build a common synchronized data frame.
This frame is used for resolving spatial relations \cite{rey2010pysal} in order to determine the potential contacts between hands and objects, similar to the method presented by Waechter at al. in \cite{7029975}.
However, we do not utilize any marker-based capturing systems for observing the task.
As mentioned above, all of the observations during the demonstration occur in 2D image space.
We assign detected objects to possible actions considering affordances modeled in ontology.\\
In the following, we show in detail how actions are associated with the robot's observations based on sequences of pre-defined constraints.
Every constraint can be evaluated to be \textit{true} or \textit{false}.
Consider the following set of constraints $C_k, k \in \{1,...,12\}$ that we apply to every data frame $f_i$ :

\begin{enumerate}
	
	\item $C_1$ : The object $o$ is classified as an active object:
	\begin{equation} \label{c1}
		g_m(o) = 1
	\end{equation}
	
	\item $C_2$ : The object $o$ is classified as a passive object:
		\begin{equation} \label{c2}
			g_m(o) = 0
		\end{equation}
	
	\item $C_3$: A set of contextual properties modeled in the ontology for active object $o_a$ contains the affordance $a$:
	\begin{equation} \label{c3}
		g_a(o_a, a) = 1
	\end{equation}
	
	\item $C_4$: A set of contextual properties modeled in the ontology for passive object $o_p$ contains the affordance $a$:
	\begin{equation} \label{c4}
		g_a(o_p, a) = 1
	\end{equation}
	
	\item $C_5$: A distance between a hand $h$ and an active object's position $p_c$ is smaller than the distance threshold $th_d$:
	\begin{equation} \label{c5}
		|p_c - h| < th_d
	\end{equation}
	
	\item $C_6$: A distance between a hand $h$ and an active object's position $p_c$ is greater than the distance threshold $th_d$:
		\begin{equation} \label{c6}
		|p_c - h| > th_d
		\end{equation}
	
	\item $C_7$: The number of data frames $j$ with valid condition $C_5$ is greater or equals the frame threshold $th_n$:
	\begin{equation} \label{c7}
		 j >= th_n 
	\end{equation}
	
	\item $C_8$: The number of data frames $j$ with valid condition $C_6$ is greater or equals the frame threshold $th_n$:
	\begin{equation} \label{c8}
		j >= th_n 
	\end{equation}
	
	\item $C_9$: Position changes for the active object as well as for the hand have been detected in two consecutive frames:
	\begin{equation} \label{c9}
		|p_{c,i} - p_{c,i+1}| = |h_i - h_{i+1}| > 0 		
	\end{equation}
	
	\item $C_{10}$: A rotation of the active object in consecutive frames is detected:
	\begin{equation} \label{c10}
		\vec{v}_i \not\equiv \vec{v}_{i+1}. 
	\end{equation}
	
	\item $C_{11}$: The active object is on the table:
	\begin{equation} \label{c11}
		t_{tl} < p_{c_a} < t_{br}
	\end{equation}
	
	\item $C_{12}$: The active object $o_a$ is located over the passive object $o_p$:
	\begin{equation} \label{c12}
	\begin{aligned}
			p_{c_a} = [x_a, y_a], \\
			p_{c_p} = [x_p, y_p], \\
			x_a = x_p, y_a < y_p	
	\end{aligned}
	\end{equation}
	\end{enumerate}
Note, in practice all equalities used in constrains ($C_{12}, C_9$) are modeled with a small $\sigma$ to allow approximate equality.

We can represent a number of basic actions like object picking,  placing, pouring and putting using this generalized constraints.
For instance, a picking action  $Pick_{o},_{h}$ is defined  as a sequence on a subset of $C$ : 
\begin{equation}
	Pick_{o},_{h} = \{C_1, C_3, C_5, C_7, C_9\}.
\end{equation}
For association of the picking action with a set of observations, the list of affordances retrieved from the semantic knowledge base (constraint $C_3$) must contain the property $pick$ (\figname\,\ref{fig:ontology}).  

For the placing action $Place_{o_k},_{h_i} \subset C$ we define the following sequence of constraints: 
\begin{equation}
	Place_{o},_{h} = \{C_1, C_3, C_6, !C_9, C_{11}\}.
\end{equation}
Analogically to the picking action, we can detect the placing action only if a correspondent property $place$ exists in the knowledge base for the active object being placed.

Once the action is defined as a sequence of constraints we store it similar to the approach presented by Waechter et al. \cite{7029975}.
Then we can reuse stored actions for building sub-sequences of constraints for association of more complex actions. 
In the following, we present the association of a pouring action.
The pouring action involves two types of objects during execution: a \textit{passive} object with property \text{accept\_pouring} retrieved through constraint $C_4$ and an \textit{active} object with property \textit{pour} from constraint $C_3$.
We denote an active object as $o_a$ and a passive object as $o_p$.   
Further, for association of the pouring action we calculate the approximate local changes $p_{o_a}$ of the active objects over time as defined in Section\,\ref{sec:approach}.
Consider the sequence of constraints for the pouring action:
\begin{equation} \label{pour}
	Pour(o_a, h, o_p) = \{ Pick_{o_a},_{h}, C_2, C_3, C_4, C_9 \} \,.
\end{equation}
Using this sequence of constraints we can infer more complex activities, for instance watering a plant: \\
\begin{equation} \label{watering}
	wateringPlant(o_a, h, o_p) = \{ Pour(o_a, h, o_p), Place(o_a, h) \},
\end{equation}  
where $o_a$ is a watering can and $o_p$ a plant.\\
Once the robot associated observations with an action it stores the sequence for later re-executions.

\subsection{Action Execution}
\label{sub:action_execution}

Resulting actions from the analysis are taken and executed sequentially. Reusable
actions have been modeled in advance and are based on a subset of actions available
on our service robot \cite{Seib2016THU}.

For grasping an object the robot executes an initial scene analysis. Object position
estimates are then transformed into a common coordinate system with the robot
manipulator. For manipulating the object we use a full body trajectory execution,
meaning that the trajectory contains movements for the robot's torso to adjust it's
height, as well as the robots arm. The trajectory is calculated by a single motion path query \cite{kuffner2000rrt}
provided by a motion planning library \cite{sucan2012open}.

In order to avoid obstacles we use the segmented plane of the tabletop segmentation
as input for a local 3D gridmap \cite{hornung2013octomap} representation of the scene.
We found this to be more error prone than using the full pointcloud as input due to
noisy measurements.







\section{Experiments}
\label{sec:experiments}

\subsubsection{Robot Description}
As an experimental platform we use two robots: service robot Lisa \cite{Memmesheimer2017R2H} and the service robot TIAGo \cite{Tiago:2016}.
Both robots are equipped with two degree of freedom (DoF) pan-tilt units with mounted RGB-D cameras on top of it. 
For voice interaction we mounted a directional microphone and the robot reports it's current action
by text to speech synthesis. Lisa performs manipulation tasks using 6 DoF Kinova Mico arm. TIAGo has an arm with 7 degrees of freedom.

\begin{figure}[t!] \centering$
  \begin{array}{cc}
    \includegraphics[height=.35\linewidth]{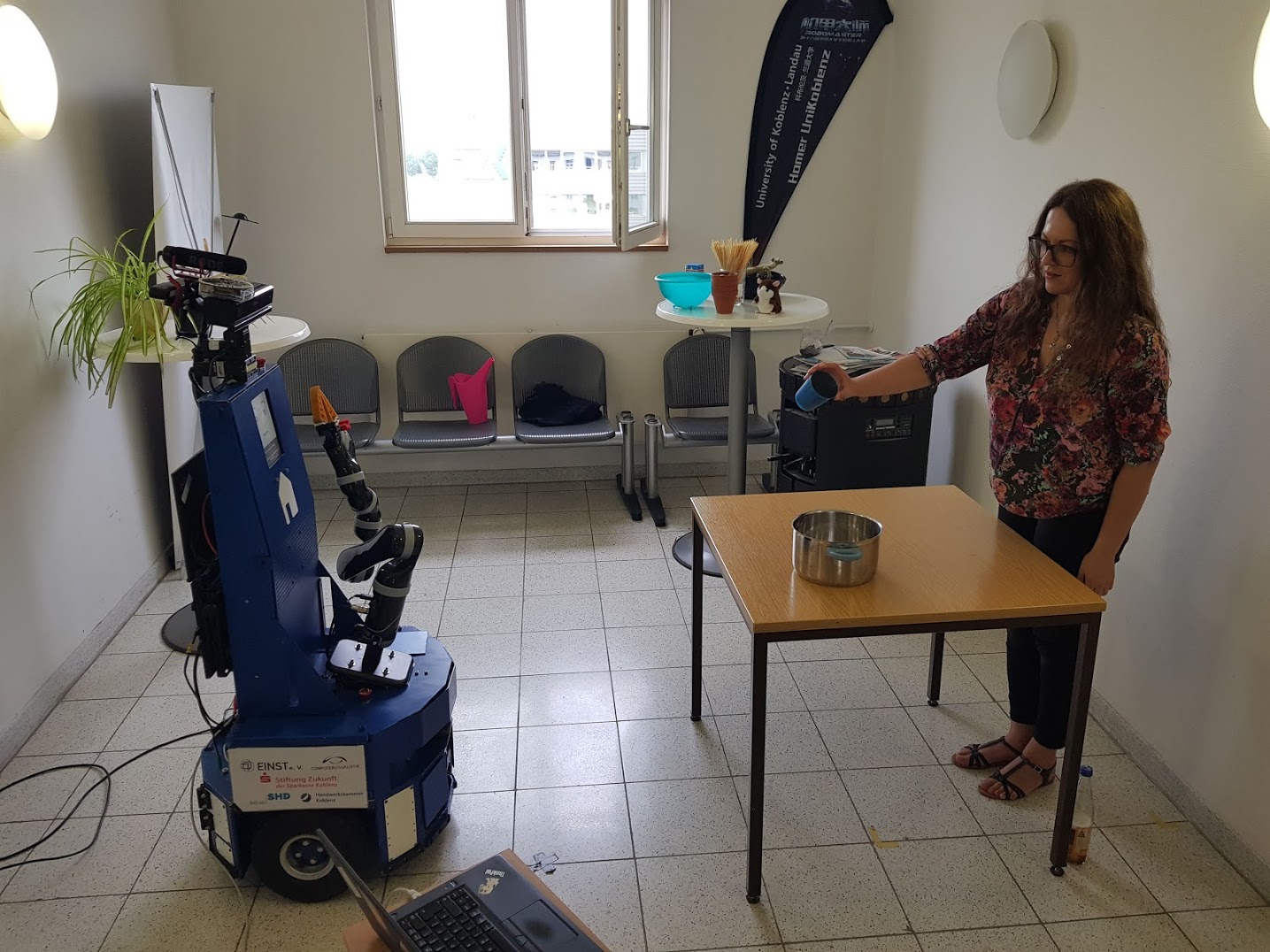} &
    \includegraphics[height=.35\linewidth]{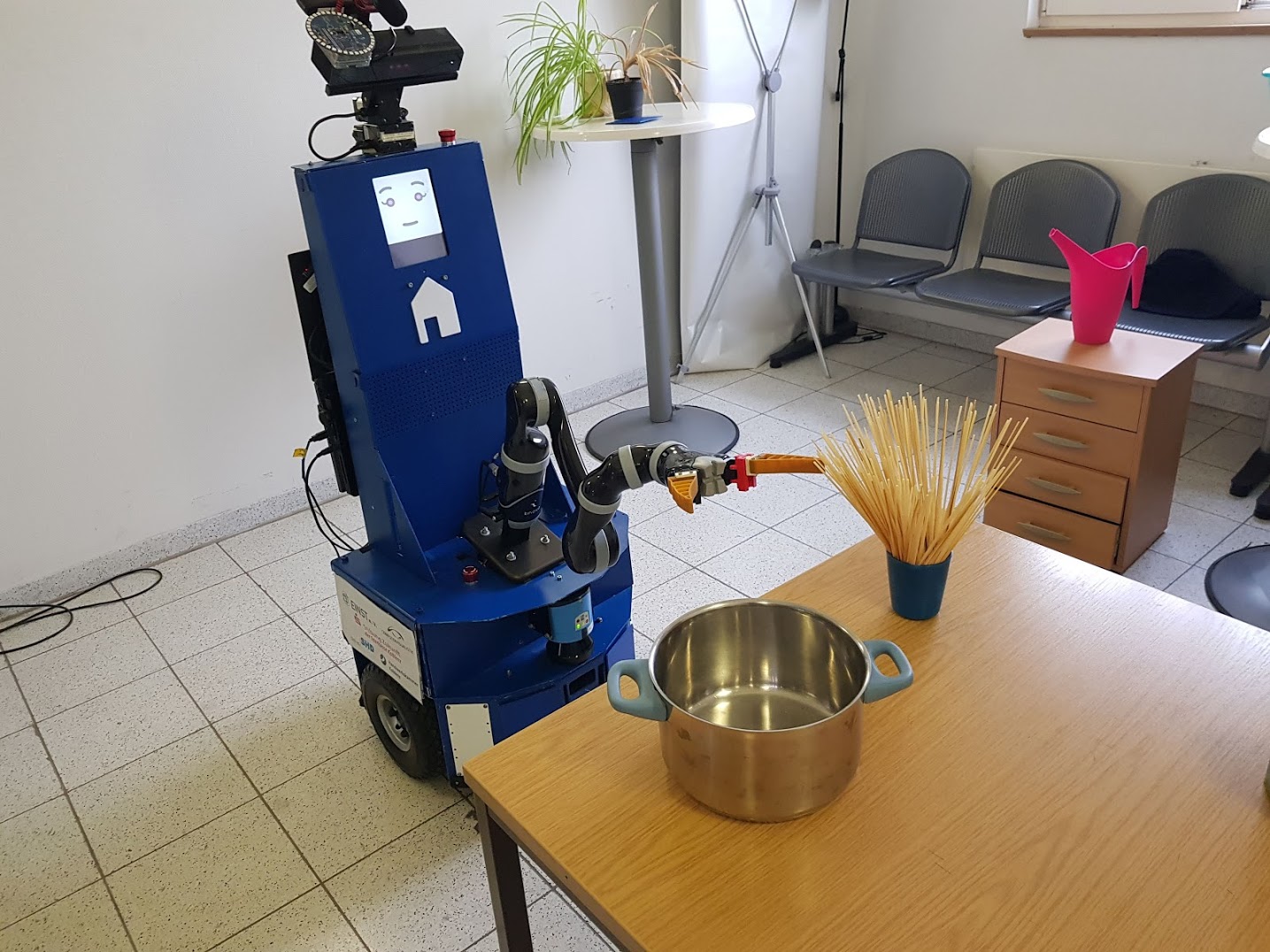} \\
  \end{array}$
  \caption{Experimental setup (left) where the robot observes a human demonstrator and executes the observed task (right)}
  \label{fig:experimental_setup}
\end{figure}

\begin{figure*}[t!]
	\centering
    \includegraphics[width=7.2in]{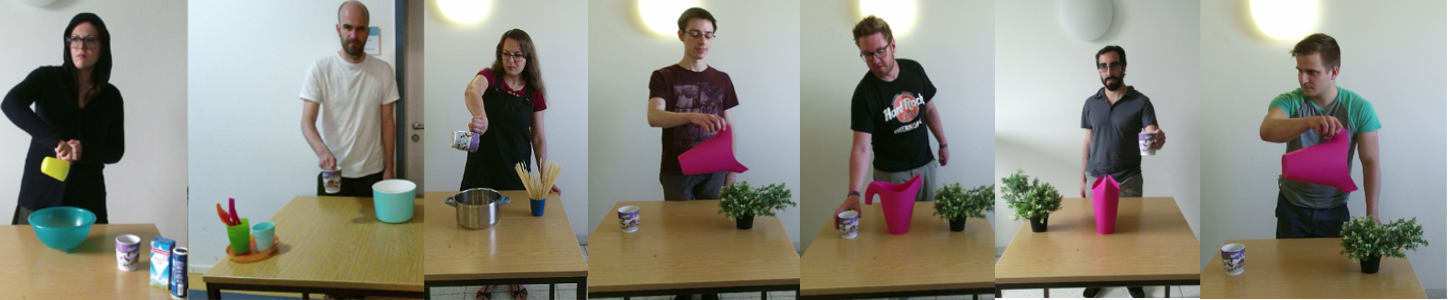}
	\caption{
	 Different human demonstrators performing actions from the robots perspective in two setups: watering a plant and preparing meal.}
	\label{fig:experiments}
\end{figure*}
\subsubsection{Setup description}
For experiments we use two different setups.
The first scenario was built within a kitchen environment with common objects used for the preparation of a meal. For the second scenario, in which we associate the action sequence to a plant watering action, we used a simple setup with a plant, a watering can and a mug located on a table.
Note that for these setups we refer to the same action sequence \ref{pour} using different sets of objects. The experimental setup are shown in \figname \ref{fig:experimental_setup}. Exemplary images of different scenarios seen from the robot perspective are shown in \figname \ref{fig:experiments}.

In both cases the robot was placed at the opposite side of the table, facing the human demonstrator. 
The used objects have been trained in order to be recognized by the robot. The object locations as well as demonstrator's location are not known by the robot in advance. 
Next to the weights for object and human body detection we modeled possible affordances of each  
object, which rely on robot actions for later execution, as it is described in \ref{sec:approach}.
\subsubsection{Description of experiments}
In order to evaluate our approach we run 16 experiments with six different demonstrators and different object sets.
The demonstrators were asked to perform following actions in a natural way: to pick a manipulatable object, to pour with it into a non-manipulatable object and then to place the object on the table.
As discussed before, we modeled such objects in our ontology as \textit{active\_objects} and \textit{passive\_objects} meaning, that they can be manipulated only with one hand. The demonstrators used only one of the hands while performing the actions. Note, for experiments with spaghetti we adapted the actions sequence $Pour(o_a, h, o_p)$ \ref{pour}. We places spaghetti in a cup for better manipulation and "{}poured" it into a pot.
Still our object recognition system was able to recognize both spaghetti and the blue cup. We applied constraints to bounding boxes which were classified as spaghetti for association of the "{}pouring " action to the observations.


After the actions have been demonstrated, we asked the robots to execute the detected actions.
We defined a successful action association if the robot could recognize basic actions correctly in a reasonable amount of time based on the given sequence of constraints.
We do not focus on action execution in this paper, because robots use pre-defined execution modules.
Instead we focus on correct interpretation of observed action sequences and a correct interpretation based on the constraints.
Videos recorded throughout the experiments are available on our project 
page\footnote{\url{https://userpages.uni-koblenz.de/~raphael/project/imitation_learning/}}. 


\subsection{Results}
\label{sub:Results}
We analyze the success rate of the action association.
The results of our experiments are presented in three tables.
The success rates of action sequence association with activities from the experimental setup is shown in Table \ref{tab:activities}.
The results indicate that our actual approach performs well on action assigning within a preparing meal scenario, but achieves 
a worse performance for watering plant activity.
Further analysis indicate that the success of action sequence association decreases when big objects are manipulated by the demonstrator (watering can). Table \ref{tab:objects} shows the performance of the system with respect to individual manipulable objects that were involved in experiments. Results depicted in Table \ref{tab:objects} prove that our approach works well on objects that are slightly bigger then the human hand (blue cup, yellow cup, mug) achieving overall performance of 97.5\% for picking and pouring, 100\% for placing. Action association that involved the object watering can tend to be less accurate, reaching a maximum success rate of 50\% on picking and pouring and 40\% on placing. This might be overcome in adding the grasping poses to the ontology as well.
 \begin{table}[t!]
 	\caption{Success rate of association action sequences to activities }
 	\begin{center}
 		\begin{tabular}{ | l | l | l |}
 			\hline
 			\textbf{Action} & Preparing meal & Watering a plant \\ \hline
	 		Pick  & 1.0		& 0.625  	\\ \hline
 			Place & 0.875 	& 0.75  	\\ \hline
 			Pour  & 0.875 	& 0.625 	\\ \hline
 		\end{tabular}
	  	\label{tab:activities}
 	\end{center}
 \end{table}
 \begin{table}[t!]
 	\caption{ Success rate of action association considering involved active objects during manipulation}
 	
 	\begin{center}
 		\begin{tabular}{ | l | l | l | l|}
 		
 			\hline
 			\textbf{Object} & \textbf{Pick} & \textbf{Pour} & \textbf{Place} \\ \hline
 			Blue cup 			& 1.0 & 1.0 & 1.0  \\ \hline
 	 		Mug 				& 1.0 & 1.0 & 1.0  \\ \hline
 	 		Pink watering can 	& 0.5 & 0.5 & 0.4  \\ \hline
 			Yellow cup 			& 1.0 & 1.0 & 1.0  \\ \hline
 			Spaghetti 			& 0.9 & 0.9 & ---  \\ \hline
 			
 		\end{tabular}
  	\label{tab:objects}
 	\end{center}
 \end{table}

We consider now constraints defined in Section \ref{sec:approach} in details. Table \ref{tab:tab2} shows success rate of applying constraints to objects used in the experiments. Note that we skipped the first four constrains as they are not dynamic and do not change during demonstrations.
Almost all constraints are recognized well with manipulable and non-manipulable objects achieving a high success rate of 100\%.
In case of using the watering can the distance from object centroid to wrist was not small enough to evaluate appropriate constraints as true.
During some demonstrations the distance between the hand and the watering can was small enough only in some frames, which we considered as noise.
Depending on the orientation of the watering can the distance to the hand was greater, then the hand object threshold $th_d$ during demonstrations. Therefore, the placing action could not be associated correctly. 

 \begin{table}[t!]
 	\caption{Success rate of action association considering constrains }
 	
 	\begin{center}
 		\begin{tabular}{| l | l | l | l | l | l |l | l|l | l | l | l | l |l | l|}
 			\hline
 			\textbf{Object/Const.} & $C_5$& $C_6$& $C_7$& $C_8$& $C_9$& $C_{10}$& $C_{11}$ & $C_{12}$ \\ \hline
 			Blue cup 				& 	 1.0 & 1.0 & 1.0 & 1.0 & 1.0 & 1.0 & 1.0 & 1.0  \\ \hline
 			Mug 					&    1.0 & 1.0 & 1.0 & 1.0 & 1.0 & 1.0 & 1.0 & 1.0  \\ \hline
 			Watering can 		    &    0.6 & 0.4 & 0.4 & 0.4 & 1.0 & 1.0 & 1.0 & 1.0  \\ \hline
 			Yellow cup 				&    1.0 & 1.0 & 1.0 & 1.0 & 1.0 & 1.0 & 1.0 & 1.0  \\ \hline
 		 	Spaghetti 				&    1.0 & 1.0 & 1.0 & 1.0 & 1.0 & 1.0 & --- & 1.0  \\ \hline
 		\end{tabular}
 		\label{tab:tab2}
 	\end{center}
 \end{table}



\section{Conclusion}
\label{sec:conclusion}

We presented an approach to imitate human actions with a domestic service
robot. The imitation is based on observations using only on-board sensors - without any augmentations of the 
environment by markers.
Convolutional pose machines (CPM) are used for estimating the demonstrator's joints during demonstration. 
Bounding boxes of objects are detected over the period of the demonstration.
The spatial relations between the demonstrator's hand
and the observed objects in 2D pixel space together with modeled affordances in a knowledge base are used to define constrains for action association.
The associated actions are then executed using a set of available actions on a domestic service robot.
Results were demonstrated on a real robot by observing actions to prepare a meal and water a plant.

For future development we plan to add a continuous object segmentation \cite{badrinarayanan2015segnet} 
to augment the current approach that is based on an initial object analysis of the scene. 
An interesting extension to the proposed approach would be to consider using constraints in 3D robot coordinate system, 
as we detected some limitations during action association using 2D image space. Another extension would be to incorporate finger detection into the observation.
Furthermore extending the current approach by pose estimations of the object or
hand will allow further reasoning about more detailed action parameters i.e. for
pouring.


\subsubsection*{Acknowledgement}

We want to thank PAL Robotics for supporting us with the award of a TIAGo 
robot in terms of the European Robotics league.

\bibliographystyle{IEEEtranM}

\bibliography{references}

\end{document}